\documentclass{article}

\usepackage[preprint]{neurips_2023}
\usepackage[utf8]{inputenc} % allow utf-8 input
\usepackage[T1]{fontenc}    % use 8-bit T1 fonts
\usepackage[hidelinks]{hyperref}       % hyperlinks
\usepackage{url}            % simple URL typesetting
\usepackage{booktabs}       % professional-quality tables
\usepackage{amsfonts}       % blackboard math symbols
\usepackage{nicefrac}       % compact symbols for 1/2, etc.
\usepackage{microtype}      % microtypography
\usepackage{xcolor}         % colors
\usepackage{tabularx}
\usepackage{graphicx}
\usepackage{multirow}
\usepackage{tikz-dependency}
\usepackage{pgfplots}
\usepackage{dashrule}
\pgfplotsset{compat=1.18}
\usepgfplotslibrary{polar}
\usepackage{pgfplotstable}
\usepackage{tkz-kiviat}
\usepackage{soul}
\usepackage{geometry}
\usepackage{ltablex}
\usepackage{pgfplots}
\usepackage{tikz}
\usepgfplotslibrary{groupplots}
\usepackage{adjustbox}
\usepackage{fontawesome5} % For additional icons
\usepackage{wrapfig}
\usepackage{enumitem}

\newcommand{\model}{\textsc{Flacuna}}

\newcommand{\instructeval}{\textsc{InstructEval}}
\newcommand{\vicuna}{\textsc{Vicuna}}
\newcommand{\flan}{\textsc{Flan}}
\newcommand{\flanmini}{\textsc{Flan-mini}}
\newcommand{\ftf}{\textsc{Flan-T5}}

\let\realcite\cite
\renewcommand{\cite}[1]{\ifx.#1.\hl{[?]}\else\realcite{#1}\fi}

\title{\model{}: Unleashing the Problem Solving Power of \textsc{Vicuna} using \textsc{Flan} Fine-Tuning}

\author{
  Deepanway Ghosal\textsuperscript{\ddag}, Yew Ken Chia\textsuperscript{\ddag}, Navonil Majumder\textsuperscript{\dag}, Soujanya Poria\textsuperscript{\ddag} \\
  \textsuperscript{\ddag} DeCLaRe Lab, Singapore University of Technology and Design, Singapore \\
  \texttt{\{deepanway\_ghosal, yewken\_chia\}@mymail.sutd.edu.sg} \\
  \texttt{\{navonil\_majumder,sporia\}@sutd.edu.sg} \\
}

\begin{document}

\maketitle
\begin{minipage}[t]{\linewidth}
  \begin{center}
    \includegraphics[width=0.5\linewidth]{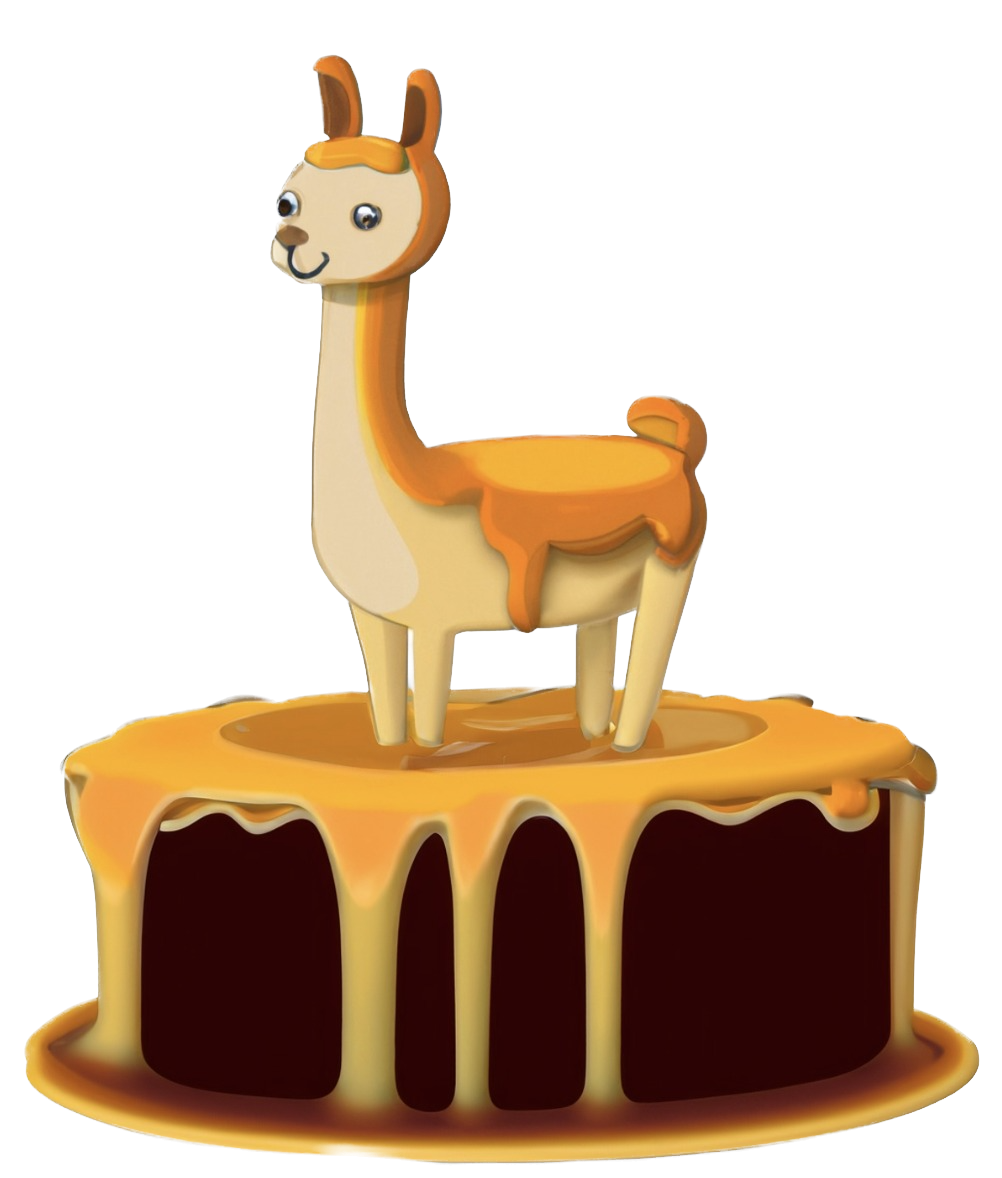}
  \end{center}
\end{minipage}
\vspace{0.3cm}
\begin{minipage}[t]{\linewidth}
  \centering
  \textsc{Code}: \url{https://github.com/declare-lab/flacuna} \\
  \textsc{Model}: \url{https://huggingface.co/declare-lab/flacuna-13b-v1.0} \\
  \flanmini{}: \url{https://huggingface.co/declare-lab/flan-mini}
\end{minipage}

\begin{abstract}
Recently, the release of \instructeval{}~\citep{instructeval} has provided valuable insights into the performance of large language models (LLMs) that utilize encoder-decoder or decoder-only architecture. Interestingly, despite being introduced four years ago, \textsc{T5}-based LLMs, such as \textsc{Flan-T5}, continue to outperform the latest decoder-based LLMs, such as \textsc{LLaMA} and \textsc{Vicuna}, on tasks that require general problem-solving skills. This performance discrepancy can be attributed to three key factors: (1) Pre-training data, (2) Backbone architecture, and (3) Instruction dataset. In this technical report, our main focus is on investigating the impact of the third factor by leveraging \textsc{Vicuna}, a large language model based on \textsc{LLaMA}, which has undergone fine-tuning on ChatGPT conversations. To achieve this objective, we fine-tuned \textsc{Vicuna} using a customized instruction dataset collection called 
\flanmini{}.
This collection includes a subset of the 
% comprehensive 
large-scale
instruction dataset known as \textsc{Flan}, as well as various code-related datasets and conversational datasets derived from ChatGPT/GPT-4. This dataset comprises a large number of tasks that demand problem-solving skills. Our experimental findings strongly indicate that the enhanced problem-solving abilities of our model, \model{}, are obtained through fine-tuning \textsc{Vicuna} on the \textsc{Flan} dataset, leading to significant improvements across numerous benchmark datasets in \instructeval{}. \model{} is publicly available at \url{https://huggingface.co/declare-lab/flacuna-13b-v1.0}.
\end{abstract}

\section{Introduction}
\label{sec:intro}

ChatGPT and its successor GPT-4 have surpassed their prior state-of-the-art models on a vast majority of the benchmarking tasks and datasets. However, to preserve privacy, natively running a 175B+ sized model like GPT-3 is beyond the capabilities of most organizations, let alone individuals. This has prompted many researchers to fine-tune manageable-sized LLMs --- from 7B to 30B on a diverse set of 
% tasks 
instruction examples
generated by ChatGPT or GPT-4. This has birthed LLMs, such as, Alpaca~\citep{alpaca} and \vicuna{}~\citep{vicuna2023} that are fine-tuned checkpoints of LLaMA~\citep{llama}. These models have attained close to ChatGPT-level performance on some specific benchmarking tasks, but overall generalization still remains elusive. Recent works like \instructeval{}~\citep{instructeval} strongly hint that the fine-tuning datasets dictate the task-specific performances. For instance, it has been observed that \textsc{Flan-T5} --- a \textsc{T5} checkpoint fine-tuned on \textsc{Flan} Collection instruction dataset --- outperforms \vicuna{} and Alpaca on tasks involving strong reasoning and problem-solving skills. This spurred us to fine-tune \vicuna{} on \textsc{Flan-mini} Collection dataset, anticipating improvement on reasoning-intensive tasks in \instructeval{}~\citep{instructeval}.

To this end, we first sample a 1M-sized instruction dataset from the 15M-sized \textsc{Flan} Collection dataset~\citep{longpre2023flan} and combined it with several other datasets comprising coding tasks and ChatGPT/GPT-4 distilled conversations. The resulting smaller dataset, \textsc{Flan-mini}, is then cast into the conversational format of \vicuna{}. 
% To make the fine-tuning parameter efficient,
To ensure a reasonable computational cost for the fine-tuning process,
we retrofit LoRA~\citep{hu2021lora} adapter into the LLaMA~\citep{llama} decoder-transformer of \vicuna{}. Following a parameter-efficient LoRA fine-tuning of the \vicuna{} checkpoint on \textsc{Flan-mini}, we obtain \model{}. As expected, \model{} outperforms \vicuna{} by a substantial margin on most benchmark datasets, especially for reasoning-intensive tasks. However, the performance of \model{} still remains below \ftf{} on the same reasoning benchmarks. This could be attributed to the 15-times smaller dataset of the instruction dataset which may contain less diverse samples. Furthermore, full fine-tuning of \vicuna{} may narrow the gap with \ftf{}.

This work overall has the following contributions:
\begin{enumerate}[leftmargin=*]
    \item Improving the problem-solving capability of \vicuna{} through parameter efficient fine-tuning on \textsc{Flan-mini}.
    \item Introducing an instruction tuning dataset, \textsc{Flan-mini}, comprising a diverse set of tasks and templates.
\end{enumerate}

\section{Training Details}
\paragraph{Preparing the \textsc{Flan-mini} Collection.}
Given the enormous size of the \textsc{Flan} Collection~\citep{longpre2023flan}, we opted to work with a carefully selected subset that maintains a high level of task diversity while reducing the overall dataset size. In Table \ref{tab:flan}, we present the specific tasks included in our subset of \textsc{Flan}, along with their respective dataset sizes. 
%\sj{Need to give details on the other datasets ie code search, code contest, apps}
%\ken{
As the public release of the \textsc{Flan} Collection does not include programming tasks, we augment the collection with existing code datasets.
Specifically, we include CodeContests \citep{Li_2022}, APPS \citep{Hendrycks2021MeasuringCC} and CodeSearchNet \citep{Husain2019CodeSearchNetCE}.
Following the data processing pipeline of \textsc{Flan} Collection, we sample a fixed number of examples from each dataset, where each example is randomly augmented with different prompt templates.
Specifically, the examples are processed with a pool of handcrafted prompt templates and may be used as zero-shot examples or grouped together with few-shot demonstrations \citep{longpre2023flan}.
%}

\begin{table*}[ht!]
  \centering
  \begin{tabular}{lcc}
    \toprule
    \textbf{Dataset Name} & \textbf{Source} & \textbf{Dataset Size} \\
    \midrule
    Flan2021 & \textsc{Flan} & 388K \\
    Public Pool of Prompts & \textsc{Flan} & 320K \\
    Natural instructions v2 & \textsc{Flan} & 200K \\
    CoT & \textsc{Flan} & 100K \\
    Code Search & \citet{husain2019codesearchnet}  & 100K \\
    Code Contest & \citet{li2022competition} & 50K \\
    Apps & \citet{hendrycksapps2021} & 50K  \\
    \midrule
    GPT4-Alpaca & GPT-4 & 52K \\
    Code-Alpaca & ChatGPT & 20K \\
    ShareGPT & ChatGPT & 60K \\
    \midrule
    Total & - & 1.34M\\
    \bottomrule
  \end{tabular}
\caption{The \textsc{Flan-mini} Collection, used to train \model{}.}
  \label{tab:flan}
\end{table*}
\paragraph{Maintaining \textsc{Vicuna's} Chatting Ability.}
\textsc{Vicuna} has demonstrated remarkable chatting ability, achieving 90\% of the performance of ChatGPT. This indicates its significant potential as an open-source alternative to closed-source large language models (LLMs) like ChatGPT. To ensure that \model{} retains \textsc{Vicuna}'s learned knowledge and chatting ability, we incorporated various ChatGPT datasets, including Alpaca~\citep{alpaca}, Code Alpaca~\citep{codealpaca}, and ShareGPT~\citep{vicuna2023}, into our \textsc{Flan} collection. Among these three datasets, \textsc{Vicuna} was originally fine-tuned using the ShareGPT dataset. The final collection was then used to train \model{}.

\paragraph{Architecture.} We employed \textsc{LoRA} in the \textsc{Vicuna} model for fine-tuning on the \textsc{Flan-mini} collection. We inserted the low-rank adapters on all the query and value projection layers, resulting in a total trainable parameter count of 6.55M, which is only around 0.05\% of the parameter count of the original 13B \textsc{Vicuna} model. The maximum input sequence length was set to 1280, and efficient training was facilitated by utilizing bf16 precision. 

\paragraph{Hyperparameter Details.} 
\model{} was trained on 4$\times$A6000 GPUs for 1 epoch. We use 16 gradient accumulation steps with a per-device batch size of 2, resulting in a  total batch size of 128. We used 3000 warm-up steps and a learning rate of 2e-5.

\section{Evaluation Tasks and Results}
\subsection{Problem Solving Evaluation}
To assess the problem-solving prowess of instructed large language models (LLMs), \instructeval{} employs a range of benchmarks encompassing real-world exams that delve into diverse topics. These benchmarks encompass complex instructions, arithmetic problems, programming challenges, and causal reasoning tasks. In order to excel in these benchmarks, models need to exhibit a profound understanding of the world, demonstrate multi-hop reasoning capabilities, showcase creativity, and employ a plethora of other cognitive skills.

\paragraph{World Knowledge.}
% The Massive Multitask Language Understanding (MMLU) \citep{mmlu} benchmark is designed to measure world knowledge and problem-solving ability in multiple subjects. 
% It evaluates models in zero-shot and few-shot settings, making it more challenging and closer to how humans are evaluated. 
% The benchmark covers 57 subjects across STEM, humanities, social sciences, and other areas, ranging in difficulty from elementary to advanced professional levels. 
The Massive Multitask Language Understanding (MMLU) benchmark, introduced in the work by \citet{mmlu}, serves as an assessment tool to gauge the problem-solving aptitude and world knowledge of language models across various subjects. It offers evaluations in both zero-shot and few-shot settings, presenting a more challenging and human-like evaluation scenario. The MMLU benchmark encompasses a comprehensive range of 57 subjects spanning STEM, humanities, social sciences, and other domains. The difficulty levels of the tasks within the benchmark vary from elementary to advanced professional levels, providing a comprehensive assessment of the model's capabilities in problem-solving and domain understanding.

\paragraph{Complex Instructions.}
The subset known as BIG-Bench Hard (BBH) comprises 23 highly demanding tasks carefully selected from the BIG-Bench benchmark \citep{srivastava2022imitation} to specifically target tasks that are considered to surpass the current capabilities of language models \citep{BBHSuzgun2022ChallengingBT}. BBH presents models with intricate instructions that require advanced skills in navigation, logical deduction, and fallacy detection.

\paragraph{Comprehension and Arithmetic.}
Discrete Reasoning Over Paragraphs (DROP) is a reading comprehension task with a mathematical focus. It challenges systems to engage in discrete reasoning by analyzing passages extracted from Wikipedia articles. In order to excel in the DROP task, a system needs to adeptly navigate references within a question and identify the appropriate sections of the provided passage. Additionally, the system must demonstrate proficiency in performing discrete operations like addition, counting, or sorting.
\paragraph{Programming.}
HumanEval serves as a problem-solving benchmark specifically designed for assessing the performance of large language models that are trained on code \citep{HumanEvalChen2021EvaluatingLL}. The benchmark comprises 164 unique programming problems, encompassing areas such as language comprehension, algorithms, and basic mathematics. Some of the problems included in HumanEval are similar in nature to straightforward software interview questions. In the evaluation process, models are assessed based on the functional correctness of the code programs they generate, with the criteria for correctness determined by the given docstrings. HumanEval provides a comprehensive evaluation framework for assessing the problem-solving capabilities of language models in a code-centric context.

\paragraph{Causality.}
The Counterfactual Reasoning Assessment (CRASS) benchmark is a novel dataset and evaluation tool developed specifically to assess the causal reasoning abilities of large language models. By employing counterfactual scenarios, CRASS tests the model's capability to identify and select appropriate causal explanations. This benchmark provides a unique and rigorous evaluation framework to gauge the causal reasoning capabilities of language models.

\subsection{Alignment to Human Values}
Noting the importance of aligning LLMs to human values, \instructeval{} incorporates the Helpful, Honest, and Harmless (HHH) benchmark \citep{askell2021general}. The benchmark showcases engaging dialogues between humans and conversational assistants, challenging the model to discern and provide the most appropriate response. It encompasses a diverse array of 61 honesty-related, 59 helpfulness-related, and 58 harmlessness-related samples, along with 43 unique instances falling within the "other" category. The inclusion of the "other" category accounts for examples that embody values not explicitly covered by honesty, helpfulness, or harmlessness.

\subsection{Writing Experiments}
For the writing experiment, we utilized the \textsc{IMPACT} dataset, which is readily available in \instructeval{}. This comprehensive dataset consists of 50 prompts across distinct categories, namely informative, professional, argumentative, and creative. Following that, ChatGPT was assigned the responsibility of scoring the models' responses in terms of relevance (Rel.) and coherence (Coh.) on a scale of 1 to 5. For more comprehensive information regarding this evaluation, we refer readers to \citet{instructeval}. 

\subsection{Results}
\paragraph{Comparative Baselines.}
As baselines, we selected \textsc{Vicuna}~\citep{zheng2023judging} and \textsc{StableVicuna}\footnote{\url{https://huggingface.co/CarperAI/stable-vicuna-13b-delta}}.

\paragraph{Few-shot Problem-solving.}
We present the results of \model{} on five datasets (see Table \ref{tab:results}) from the \instructeval{} benchmark, focusing on problem-solving tasks. In 4 out of 5 tasks, \model{} outperformed \textsc{Vicuna}, showing an average performance improvement of 5.6 points over the LLaMA backbone. However, it performed slightly worse on code-related problem-solving tasks in the HumanEval dataset, with a margin of 0.6 points. Overall, the improvement in \model{} compared to \textsc{Vicuna} is 5.1 points averaged over the five tasks.

\begin{table*}[!t]
    \centering
    %\small
    \resizebox{\textwidth}{!}{
    \begin{tabular}{lccccccccccccc}
    \toprule
    \multirow{2}{*}{{\textbf{Model}}} & \multirow{2}{*}{{\textbf{Size}}} 
    & \multicolumn{2}{c}{\textbf{MMLU (5-shot)}} 
    & \multicolumn{2}{c}{\textbf{BBH (3-shot)}} 
    & \multicolumn{2}{c}{\textbf{DROP$\star$ (3-shot)}} 
    & \multicolumn{2}{c}{\textbf{CRASS (3-shot)}} 
    & \multicolumn{2}{c}{\textbf{HumanEval (0-shot)}} 
    & \multicolumn{2}{c}{\textbf{Avg.}}  
    \\
    \cmidrule(lr){3-4} 
    \cmidrule(lr){5-6} 
    \cmidrule(lr){7-8} 
    \cmidrule(lr){9-10} 
    \cmidrule(lr){11-12} 
    \cmidrule(lr){13-14} 
     &  & Perf. & $\Delta$ & Perf. & $\Delta$ & Perf. & $\Delta$ & Perf. & $\Delta$ & Perf. & $\Delta$ & Perf. & $\Delta$ \\
    \midrule
    % Human & - & 89.8\footnote{Estimation of human expert-performance \citep{mmlu}} & \\
    GPT-4 & - & 86.4 & - & - & - & 80.9 & - & - & - & 67.0 & - & - & - \\
    ChatGPT & - & 70.0 & - & 49.5 & - & 64.1 & - & 90.5 & - & 48.1 & - & 64.5 & - \\
    \midrule
    Flan-UL2 & 20B & 55.0 & - & 44.7 & - & 64.3 & - & 94.2 & - & 0.0 & - & 51.6 & -  \\
    Alpaca-Lora & 30B & 58.4 & +0.6 & 41.3 & +2.0 & 45.1 & -0.3 & 79.2 & +10.6 & 18.9 & +4.9 & 48.6 & +3.6  \\
    OpenAssistant & 30B & 56.9 & -0.9 & 39.2 & -0.1 & 46.0 & +0.6 & 67.2 & +1.4 & 23.1 & +9.1 & 46.5 &  +1.5 \\
    OPT-IML & 30B & 38.6 & +11.3 & 31.3 & +3.0 & 47.5 & +28.0 & 67.2 & +32.5 & 9.1 & +7.9 & 38.7 & +16.5 \\
    \midrule
    Flan-T5 & 11B & 54.5 & +29.3 & 43.9 & +13.6 & 67.2 & +49.7 & 88.3 & +54.7 & 0.0 & +0.0 & 50.8 & +29.5 \\
    Flan-Alpaca & 11B & 50.9 & +25.7 & 23.3 & -7.0 & 62.3 & +44.8 & 90.2 & +56.6 & 0.0 & +0.0 & 45.3 & +24.0 \\
    Dolly V2 & 12B & 25.6 & -1.3 & 29.7 & +0.2 & 16.6 & -0.5 & 35.8 & +1.1 & 8.5 & -0.6 & 23.2 & -0.7 \\
    \midrule
    Flan-T5 & 3B & 49.2 & +25.9 & 40.2 & +15.9 & 56.3 & +43.7 & 91.2 & +60.2 & 0.0 & +0.0 & 47.4 & +29.2 \\
    ChatGLM & 6B & 36.1 & - & 31.3 & - & 44.2 & - & 51.1 & - & 3.1 & - & 33.2 & -  \\
    % Alpaca-Lora & 7B & 35.6 & +0.4 & 30.7 & +0.2 & 27.5 & -0.1 & 45.6 & +11.7 & 15.9 & +5.6 & 31.1 & +3.5 \\
    Mosaic-Chat & 7B & 37.1 & +1.9 & 32.0 & +1.1 & 20.2 & -7.4 & 47.5 & +13.6 & 17.7 & +7.4 & 30.9 & +3.3 \\
    \midrule
    %\textsc{Vicuna-1.0} & 13B & 49.7 & +3.5 & 37.1 & +0.0 & 32.9 & -2.4 & 60.9 & +2.1 & 15.2 & +1.8 & 39.2 & +1.0 \\
    \textsc{StableVicuna} & 13B & 49.2 & +3.0 & 37.5 & +0.4 & 34.3 & -1.0 & 67.5 & +8.7 & 15.9 & +2.5 & 40.9 & +2.7 \\
    \textsc{Vicuna} & 13B & 50.6 & +4.5 & 37.6 & +0.5 & 32.6 & -3.0 & 60.9 & +2.1 & 11.6 & -1.8 & 38.7 & +0.5 \\
    \model{} & 13B & 51.1 & +5.0 & 39.3 & +2.2 & 43.6 & +8.0 & 74.1 & +15.3 & 11.0 & -2.4 & 43.8 & +5.6 \\
    \bottomrule
    \end{tabular}
    }
    \caption{Evaluation results for problem-solving benchmarks. 
    We denote the original performance across the benchmarks as Perf., while $\Delta$ denotes the change in performance compared to the corresponding foundation LLMs. $\star$ indicates that DROP is a held-in dataset.}
    \label{tab:results}
\end{table*}
\begin{table*}[!t]
\centering
%\small
\begin{tabular}{lcccc}
\toprule
\textbf{Model} & \textbf{Size} & \textbf{MMLU (0-shot)} & \textbf{BBH (0-shot)} & \textbf{CRASS (0-shot)} \\
\midrule
Flan-UL2 & 20B & 54.4 & 34.9 & - \\
OpenAssistant & 30B & 52.0 & 33.4 & - \\
OPT IML & 30B & 41.3 & 17.4 & - \\
\midrule
TK-Instruct & 11B & 39.4 & 17.1 & - \\
Flan-T5-XXL & 11B & 54.1 & 39.5 & - \\
\midrule
Dolly V2 & 12B & 25.4 & 22.3 & - \\
\midrule
\textsc{StableVicuna} & 13B & 47.5  & 18.5  & 64.2 \\
\textsc{Vicuna} & 13B & 48.3  & 28.3 & 65.7 \\
\model{} & 13B & 49.4  & 32.5 & 67.9 \\
% Llama-7B & 7B & 42.1 (-0.4) & 44.4 (+6.8) & 54.1 (-8.9) & 44.2 (+7.0) \\
% Wombat-7B-GPT-4 & 7B & 39.2 (-4.3) & 60.2 (-11.9) & 48.0 (-3.8) & 53.7 (+2.1) \\
% Alpaca & 7B & 49.3 & 60.3 & 51.5 & 50.2 & 52.8 \\
\bottomrule
\end{tabular}
\caption{0-shot problem-solving evaluation of \model{} and other baseline models.}
\label{tab:0-shot}
\end{table*}

Out of the five problem-solving datasets, one of them, DROP, is categorized as a held-in dataset. It is a part of our \textsc{Flan} collection and was utilized for training \model{}. As a result, we observed a significant performance boost of 11 points compared to \textsc{Vicuna}. The remaining datasets are considered held out.

\paragraph{0-shot Problem-solving.}
We conducted a 0-shot performance evaluation of \model{} and compared it against both \textsc{Vicuna} and \textsc{StableVicuna}. The results presented in Table \ref{tab:0-shot} demonstrate a noteworthy performance leap by \model{} compared to its competitors. This improvement can be attributed to the training of \model{} on the high-quality \textsc{Flan} instruction dataset.

\paragraph{HHH Evaluation.}
We conducted a further evaluation using BBH's HHH evaluation dataset (see Table \ref{hhh}), where \model{} exhibited an impressive 11\% improvement over \textsc{Vicuna}. Notably, our instruction dataset collection aimed to enhance \textsc{Vicuna}'s problem-solving abilities, but it also had a positive impact on its HHH performance. This observation aligns with the experience of \textsc{Flan-T5}, which achieved a 24.2\% performance improvement over its \textsc{T5} backbone after fine-tuning on \textsc{Flan}.

\begin{table*}[ht!]
\centering
%\small
\begin{tabular}{lccccccc}
\toprule
\textbf{Model} & \textbf{Size} & \textbf{Harmlessness} & \textbf{Helpfulness} & \textbf{Honesty} & \textbf{Other} & \textbf{Avg.} & \textbf{$\Delta$} Avg. \\
\midrule
ChatGPT & - & 90.7 & 91.2 & 78.1 & 86.3 & 86.6 & -\\
\midrule
% Flan-UL2 & 20B & 78.2 (+1.1) & 71.8 (+5.0) & 72.7 (-4.4) & 89.9 (-5.4) \\
% \midrule
Flan-Alpaca & 11B & 74.2 & 81.4  & 77.4 & 83.4 & 79.1 & +26.6 \\
Flan-T5 & 11B & 75.9 & 75.3 & 75.1 & 79.6 & 76.7 & +24.2\\
Tk-Instruct & 11B & 70.1 & 54.8 & 62.3 & 76.0 & 65.8 & +13.3\\
% Koala & 13B & 57.2 & 59.6 & 57.0 & 69.7 & 60.9 & +8.4 \\
T5 & 11B & 46.4 & 54.8 & 58.1 & 50.7 & 52.5 & - \\
\midrule
Alpaca & 13B & 49.7 & 51.2 & 51.8 & 45.5 & 49.5 & -12.3 \\
LLaMA & 13B & 57.2 & 61.0 & 57.0 & 72.0 & 61.8 & -\\
\midrule
Dolly V2 & 12B & 51.7 & 59.9 & 47.0 & 58.1 & 54.2 & +9.1 \\
Pythia & 12B & 41.3 & 46.1 & 43.6 & 49.3 & 45.1 & - \\
\midrule
\textsc{StableVicuna} & 13B & 61.7  & 67.2  & 57.1 & 79.1 & 66.3 & +4.5  \\
\textsc{Vicuna} & 13B & 62.0  & 66.1 & 52.4 & 74.4 & 63.7 & +1.9\\
\model{} & 13B & 72.4  & 71.2 & 70.5 & 83.7 & 74.5 & +12.6\\
% Llama-7B & 7B & 42.1 (-0.4) & 44.4 (+6.8) & 54.1 (-8.9) & 44.2 (+7.0) \\
% Wombat-7B-GPT-4 & 7B & 39.2 (-4.3) & 60.2 (-11.9) & 48.0 (-3.8) & 53.7 (+2.1) \\
% Alpaca & 7B & 49.3 & 60.3 & 51.5 & 50.2 & 52.8 \\
\bottomrule
\end{tabular}
\caption{Evaluation results for alignment to human values on the honesty, helpfulness, and harmlessness (HHH) benchmark. Avg. denotes the average performance, while $\Delta$ Avg. denotes the average improvement compared to the corresponding foundation model.}
\label{hhh}
\end{table*}
\begin{table*}[!t]
    \centering
    %\small
    \resizebox{\textwidth}{!}{
    \begin{tabular}{lccccccccccc}
    \toprule
    \multirow{2}{*}{{\textbf{Model}}} & \multirow{2}{*}{{\textbf{Size}}} 
    & \multicolumn{2}{c}{\textbf{Informative}} 
    & \multicolumn{2}{c}{\textbf{Professional}} 
    & \multicolumn{2}{c}{\textbf{Argumentative}} 
    & \multicolumn{2}{c}{\textbf{Creative}} 
    & \multicolumn{2}{c}{\textbf{Avg.}} 
    \\ 
    \cmidrule(lr){3-4}
    \cmidrule(lr){5-6}
    \cmidrule(lr){7-8}
    \cmidrule(lr){9-10}
    \cmidrule(lr){11-12}
    & & Rel. & Coh. & Rel. & Coh. & Rel. & Coh. & Rel. & Coh. & Rel. & Coh. \\
    \midrule
    ChatGPT & - & 3.34 & 3.98 & 3.88 & 3.96 & 3.96 & 3.82 & 3.92 & 3.94 & 3.78 & 3.93 \\
    Flan-Alpaca & 11B & 3.56 & 3.46 & 3.54 & 3.70 & 3.22 & 3.28 & 3.70 & 3.40 & 3.51 & 3.46 \\
    Flan-T5 & 11B & 2.64 & 3.24 & 2.62 & 3.22 & 2.54 & 3.40 & 2.50 & 2.72 & 2.58 & 3.15 \\
    Dolly-V2 & 12B & 3.54 & 3.64 & 2.96 & 3.74 & 3.66 & 3.20 & 3.02 & 3.18 & 3.30 & 3.44 \\
    \midrule
    \textsc{StableVicuna} & 13B & 3.54 & 3.64 & 2.96 & 3.74 & 3.30 & 3.20 & 3.02 & 3.18 & 3.21 & 3.44 \\
    \textsc{Vicuna} & 13B & 3.60	& 3.96	& 3.74	& 3.82	& 3.82	& 3.56	& 3.82	& 3.92	& 3.75	& 3.82\\
    \model{} & 13B & 3.02 & 3.42 & 3.48	& 3.52 & 3.38	& 3.02 & 3.92	& 3.80 & 3.45 & 3.44 \\
    \bottomrule
    \end{tabular}
    }
    \caption{Evaluation results for writing-based tasks.
    }
    \label{tab:subjective}
\end{table*}

\paragraph{Writing Evaluation.}
While \model{} primarily excels in problem-solving tasks, we made efforts to maintain the impressive writing and chatting ability of \textsc{Vicuna}. To achieve this, we incorporated conversational datasets generated by GPT-4, such as GPT-4-Alpaca and ShareGPT, into the \textsc{Flan-mini} collection. However, despite these efforts, we observed certain issues in \model{}'s 
writing
performance. In some cases, it generates code snippets in response to prompts that are unrelated to coding. We attribute this behavior to the significant data imbalance, where the conversational dataset constitutes only 8.2\% of the entire data mixture. Prompt engineering techniques can help rectify such issues.

We discovered that \model{} generates responses of reasonable quality when provided with the following template: \texttt{```A chat between a curious user and an artificial intelligence assistant. The assistant gives helpful, detailed, and polite answers to the user's questions. USER: {definition of the task}./n/n
{question}/n
Output: ASSISTANT: '''}. This finding suggests that with the appropriate prompts, we can improve \model{}'s chatting performance. 

However, upon careful examination of the generated samples, it becomes apparent that \model{} does not outperform \vicuna{} as 
% a chat model.
a writing assistant.
This observation is reinforced by the evaluation of the generated responses to the prompts in the IMPACT dataset using ChatGPT, as depicted in Table \ref{tab:subjective}. ChatGPT consistently ranks \vicuna{}'s responses significantly higher than those of \model{}.

The subpar performance of \model{} in 
% chat 
writing-based
scenarios can be attributed to several factors. Firstly, the disproportionate 
% presence
scarcity
of conversational datasets in \flan{} may have contributed to this outcome. 
Additionally, parameter-efficient tuning methods such as \textsc{LoRA} may limit the effectiveness of the model in learning both problem-solving and general writing abilities. Hence, we may explore other efficient training methods for LLMs in the future \citep{lv2023parameter}.
% Additionally, the inefficiency of \textsc{LoRA} in effectively learning both chat-based interactions and problem-solving tasks could also be a limiting factor.

An example of the prompt and \model{}'s response is shown below.

%\hl{Give table and compare to Vicuna.}
\keepXColumns
\begin{tabularx}{\linewidth}{X}
\toprule
% id 10
\textbf{Prompt:} \texttt{A chat between a curious user and an artificial intelligence assistant. The assistant gives helpful, detailed, and polite answers to the user's questions. USER: You are tasked to demonstrate your writing skills in professional or work settings for the following question.
Write a user manual for a household appliance, providing step-by-step instructions for its operation, maintenance, and troubleshooting, using clear and simple language.
Output: ASSISTANT: } 
\\
\\
\textbf{\model{}'s response:} \\

\begin{minipage}[c]{\textwidth}
\begin{quote}
1. Introduction

Welcome to the user manual for the XYZ household appliance. This appliance is designed to make your life easier and more convenient. With this appliance, you can prepare your meals quickly and efficiently.

2. Operation

To operate the appliance, follow these steps:

a. Turn on the appliance by pressing the power button.
b. Select the desired cooking mode by using the mode selector.
c. Place the food inside the appliance.
d. Close the lid and press the start button.
e. Wait for the appliance to complete the cooking process.

3. Maintenance

To keep your appliance in good working condition, follow these steps:

a. Clean the appliance after each use.
b. Check the appliance for any signs of damage or wear and tear.
c. If you notice any issues, contact the manufacturer or a certified repair technician.

4. Troubleshooting

If you encounter any issues with your appliance, try the following troubleshooting steps:

a. Check the power cord and plug to ensure they are securely connected.
b. Check the fuse to ensure it is not blown.
c. Check the appliance manual for troubleshooting tips.
d. If the issue persists, contact the manufacturer or a certified repair technician.

5. Safety Precautions

Always follow these safety precautions when using the appliance:

a. Keep the appliance clean and free of debris.
b. Do not use the appliance if it is damaged or showing signs of wear and tear.
c. Do not leave the appliance unattended while it is in operation.
d. Always use the appliance according to the manufacturer's instructions.

6. Conclusion

Congratulations on your new XYZ household appliance! With proper use and maintenance, your appliance will provide you with years of reliable service. If you have any questions or concerns, please refer to the appliance manual or contact the manufacturer.
\end{quote}
\end{minipage}\\  \\
\bottomrule
\end{tabularx}

\section{Limitations and Future Work}
\label{sec:limit}
Despite the promising advancements of \model{} compared to \textsc{Vicuna}, we have identified some issues that require addressing:

\begin{itemize}[leftmargin=*]
\item If \model{} is asked to provide descriptive answers to questions like ``Present arguments for or against lowering the age bar for drinking,'' \model{} \textbf{generates code snippets instead}. This behavior could be attributed to its \textbf{imperfect understanding of instructions or a tendency to hallucinate}.
\item \model{} is still \textbf{significantly behind \textsc{Flan-T5}} in terms of problem-solving abilities.
\item Surprisingly, \model{} exhibits \textbf{inferior performance compared to both \textsc{LLaMA} and \textsc{Vicuna} on coding-related problems}. This outcome is unexpected, considering that we incorporated numerous coding problem-solving datasets into our instruction tuning collection.
\item \model{} is \textbf{trained with a maximum input sequence length of 1280} which limits its ability to comprehend longer input sequences.
\end{itemize}
To address these limitations and known issues, we can explore the following steps:

\begin{itemize}[leftmargin=*]
\item Based on previous studies, it has been observed that LoRA performs better with larger models~\citep{instructeval}, such as those with 30B or 65B parameters, and excels in task-specific settings. Therefore, in future work, we could enhance \model{} by \textbf{fully fine-tuning \vicuna{}, without LoRA}, particularly on the \textsc{Flan} collection. Another future work is to train \model{} on longer token length.
\item We can \textbf{incorporate the original \textsc{Flan} collection into the training process}, as it is fifteen times larger than the instruction dataset we used in this study. \textsc{Flan-T5} underwent training on this extensive collection, which resulted in remarkable problem-solving performance.
\item The chatting or writing performance of \model{} could be improved by \textbf{incorporating larger conversational datasets in \flanmini{}} and subsequently training \model{} on it. 
\end{itemize}
\bibliography{custom}

\begin{thebibliography}{20}
\providecommand{\natexlab}[1]{#1}
\providecommand{\url}[1]{\texttt{#1}}
\expandafter\ifx\csname urlstyle\endcsname\relax
  \providecommand{\doi}[1]{doi: #1}\else
  \providecommand{\doi}{doi: \begingroup \urlstyle{rm}\Url}\fi

\bibitem[Chia et~al.(2023)Chia, Hong, Bing, and Poria]{instructeval}
Yew~Ken Chia, Pengfei Hong, Lidong Bing, and Soujanya Poria.
\newblock Instructeval: Towards holistic evaluation of instruction-tuned large
  language models, 2023.

\bibitem[Taori et~al.(2023)Taori, Gulrajani, Zhang, Dubois, Li, Guestrin,
  Liang, and Hashimoto]{alpaca}
Rohan Taori, Ishaan Gulrajani, Tianyi Zhang, Yann Dubois, Xuechen Li, Carlos
  Guestrin, Percy Liang, and Tatsunori~B. Hashimoto.
\newblock Stanford alpaca: An instruction-following llama model, 2023.
\newblock URL \url{https://github.com/tatsu-lab/stanford_alpaca}.

\bibitem[Chiang et~al.(2023)Chiang, Li, Lin, Sheng, Wu, Zhang, Zheng, Zhuang,
  Zhuang, Gonzalez, Stoica, and Xing]{vicuna2023}
Wei-Lin Chiang, Zhuohan Li, Zi~Lin, Ying Sheng, Zhanghao Wu, Hao Zhang, Lianmin
  Zheng, Siyuan Zhuang, Yonghao Zhuang, Joseph~E. Gonzalez, Ion Stoica, and
  Eric~P. Xing.
\newblock Vicuna: An open-source chatbot impressing gpt-4 with 90\%* chatgpt
  quality, March 2023.
\newblock URL \url{https://vicuna.lmsys.org}.

\bibitem[Touvron et~al.(2023)Touvron, Lavril, Izacard, Martinet, Lachaux,
  Lacroix, Rozi{\`e}re, Goyal, Hambro, Azhar, Rodriguez, Joulin, Grave, and
  Lample]{llama}
Hugo Touvron, Thibaut Lavril, Gautier Izacard, Xavier Martinet, Marie-Anne
  Lachaux, Timoth{\'e}e Lacroix, Baptiste Rozi{\`e}re, Naman Goyal, Eric
  Hambro, Faisal Azhar, Aur'elien Rodriguez, Armand Joulin, Edouard Grave, and
  Guillaume Lample.
\newblock Llama: Open and efficient foundation language models.
\newblock \emph{ArXiv}, abs/2302.13971, 2023.

\bibitem[Longpre et~al.(2023)Longpre, Hou, Vu, Webson, Chung, Tay, Zhou, Le,
  Zoph, Wei, et~al.]{longpre2023flan}
Shayne Longpre, Le~Hou, Tu~Vu, Albert Webson, Hyung~Won Chung, Yi~Tay, Denny
  Zhou, Quoc~V Le, Barret Zoph, Jason Wei, et~al.
\newblock The flan collection: Designing data and methods for effective
  instruction tuning.
\newblock \emph{arXiv preprint arXiv:2301.13688}, 2023.

\bibitem[Hu et~al.(2021)Hu, Shen, Wallis, Allen-Zhu, Li, Wang, Wang, and
  Chen]{hu2021lora}
Edward~J Hu, Yelong Shen, Phillip Wallis, Zeyuan Allen-Zhu, Yuanzhi Li, Shean
  Wang, Lu~Wang, and Weizhu Chen.
\newblock Lora: Low-rank adaptation of large language models.
\newblock \emph{arXiv preprint arXiv:2106.09685}, 2021.

\bibitem[Li et~al.(2022{\natexlab{a}})Li, Choi, Chung, Kushman, Schrittwieser,
  Leblond, Eccles, Keeling, Gimeno, Lago, Hubert, Choy, de~Masson~d'Autume,
  Babuschkin, Chen, Huang, Welbl, Gowal, Cherepanov, Molloy, Mankowitz, Robson,
  Kohli, de~Freitas, Kavukcuoglu, and Vinyals]{Li_2022}
Yujia Li, David Choi, Junyoung Chung, Nate Kushman, Julian Schrittwieser,
  R{\'{e} }mi Leblond, Tom Eccles, James Keeling, Felix Gimeno, Agustin~Dal
  Lago, Thomas Hubert, Peter Choy, Cyprien de~Masson~d'Autume, Igor Babuschkin,
  Xinyun Chen, Po-Sen Huang, Johannes Welbl, Sven Gowal, Alexey Cherepanov,
  James Molloy, Daniel~J. Mankowitz, Esme~Sutherland Robson, Pushmeet Kohli,
  Nando de~Freitas, Koray Kavukcuoglu, and Oriol Vinyals.
\newblock Competition-level code generation with {AlphaCode}.
\newblock \emph{Science}, 378\penalty0 (6624):\penalty0 1092--1097, dec
  2022{\natexlab{a}}.
\newblock \doi{10.1126/science.abq1158}.
\newblock URL \url{https://doi.org/10.1126%2Fscience.abq1158}.

\bibitem[Hendrycks et~al.(2021{\natexlab{a}})Hendrycks, Basart, Kadavath,
  Mazeika, Arora, Guo, Burns, Puranik, He, Song, and
  Steinhardt]{Hendrycks2021MeasuringCC}
Dan Hendrycks, Steven Basart, Saurav Kadavath, Mantas Mazeika, Akul Arora,
  Ethan Guo, Collin Burns, Samir Puranik, Horace He, Dawn~Xiaodong Song, and
  Jacob Steinhardt.
\newblock Measuring coding challenge competence with apps.
\newblock \emph{ArXiv}, abs/2105.09938, 2021{\natexlab{a}}.

\bibitem[Husain et~al.(2019{\natexlab{a}})Husain, Wu, Gazit, Allamanis, and
  Brockschmidt]{Husain2019CodeSearchNetCE}
Hamel Husain, Hongqi Wu, Tiferet Gazit, Miltiadis Allamanis, and Marc
  Brockschmidt.
\newblock Codesearchnet challenge: Evaluating the state of semantic code
  search.
\newblock \emph{ArXiv}, abs/1909.09436, 2019{\natexlab{a}}.

\bibitem[Husain et~al.(2019{\natexlab{b}})Husain, Wu, Gazit, Allamanis, and
  Brockschmidt]{husain2019codesearchnet}
Hamel Husain, Ho-Hsiang Wu, Tiferet Gazit, Miltiadis Allamanis, and Marc
  Brockschmidt.
\newblock {CodeSearchNet} challenge: Evaluating the state of semantic code
  search.
\newblock \emph{arXiv preprint arXiv:1909.09436}, 2019{\natexlab{b}}.

\bibitem[Li et~al.(2022{\natexlab{b}})Li, Choi, Chung, Kushman, Schrittwieser,
  Leblond, Eccles, Keeling, Gimeno, Dal~Lago, Hubert, Choy, de~Masson~d'Autume,
  Babuschkin, Chen, Huang, Welbl, Gowal, Cherepanov, Molloy, Mankowitz,
  Sutherland~Robson, Kohli, de~Freitas, Kavukcuoglu, and
  Vinyals]{li2022competition}
Yujia Li, David Choi, Junyoung Chung, Nate Kushman, Julian Schrittwieser,
  R{\'e}mi Leblond, Tom Eccles, James Keeling, Felix Gimeno, Agustin Dal~Lago,
  Thomas Hubert, Peter Choy, Cyprien de~Masson~d'Autume, Igor Babuschkin,
  Xinyun Chen, Po-Sen Huang, Johannes Welbl, Sven Gowal, Alexey Cherepanov,
  James Molloy, Daniel Mankowitz, Esme Sutherland~Robson, Pushmeet Kohli, Nando
  de~Freitas, Koray Kavukcuoglu, and Oriol Vinyals.
\newblock Competition-level code generation with alphacode.
\newblock \emph{arXiv preprint arXiv:2203.07814}, 2022{\natexlab{b}}.

\bibitem[Hendrycks et~al.(2021{\natexlab{b}})Hendrycks, Basart, Kadavath,
  Mazeika, Arora, Guo, Burns, Puranik, He, Song, and
  Steinhardt]{hendrycksapps2021}
Dan Hendrycks, Steven Basart, Saurav Kadavath, Mantas Mazeika, Akul Arora,
  Ethan Guo, Collin Burns, Samir Puranik, Horace He, Dawn Song, and Jacob
  Steinhardt.
\newblock Measuring coding challenge competence with apps.
\newblock \emph{NeurIPS}, 2021{\natexlab{b}}.

\bibitem[Chaudhary(2023)]{codealpaca}
Sahil Chaudhary.
\newblock Code alpaca: An instruction-following llama model for code
  generation.
\newblock \url{https://github.com/sahil280114/codealpaca}, 2023.

\bibitem[Hendrycks et~al.(2021{\natexlab{c}})Hendrycks, Burns, Basart, Zou,
  Mazeika, Song, and Steinhardt]{mmlu}
Dan Hendrycks, Collin Burns, Steven Basart, Andy Zou, Mantas Mazeika, Dawn
  Song, and Jacob Steinhardt.
\newblock Measuring massive multitask language understanding.
\newblock In \emph{International Conference on Learning Representations},
  2021{\natexlab{c}}.
\newblock URL \url{https://openreview.net/forum?id=d7KBjmI3GmQ}.

\bibitem[Srivastava et~al.(2022)Srivastava, Rastogi, Rao, Shoeb, Abid, Fisch,
  Brown, Santoro, Gupta, Garriga-Alonso, Kluska, Lewkowycz, Agarwal, Power,
  Ray, Warstadt, Kocurek, Safaya, Tazarv, Xiang, Parrish, Nie, Hussain, Askell,
  Dsouza, Slone, Rahane, Iyer, Andreassen, Madotto, Santilli, Stuhlmüller,
  Dai, La, Lampinen, Zou, Jiang, Chen, Vuong, Gupta, Gottardi, Norelli,
  Venkatesh, Gholamidavoodi, Tabassum, Menezes, Kirubarajan, Mullokandov,
  Sabharwal, Herrick, Efrat, Erdem, Karakaş, Roberts, Loe, Zoph, Bojanowski,
  Özyurt, Hedayatnia, Neyshabur, Inden, Stein, Ekmekci, Lin, Howald, Diao,
  Dour, Stinson, Argueta, Ramírez, Singh, Rathkopf, Meng, Baral, Wu,
  Callison-Burch, Waites, Voigt, Manning, Potts, Ramirez, Rivera, Siro, Raffel,
  Ashcraft, Garbacea, Sileo, Garrette, Hendrycks, Kilman, Roth, Freeman,
  Khashabi, Levy, González, Perszyk, Hernandez, Chen, Ippolito, Gilboa, Dohan,
  Drakard, Jurgens, Datta, Ganguli, Emelin, Kleyko, Yuret, Chen, Tam, Hupkes,
  Misra, Buzan, Mollo, Yang, Lee, Shutova, Cubuk, Segal, Hagerman, Barnes,
  Donoway, Pavlick, Rodola, Lam, Chu, Tang, Erdem, Chang, Chi, Dyer, Jerzak,
  Kim, Manyasi, Zheltonozhskii, Xia, Siar, Martínez-Plumed, Happé, Chollet,
  Rong, Mishra, Winata, de~Melo, Kruszewski, Parascandolo, Mariani, Wang,
  Jaimovitch-López, Betz, Gur-Ari, Galijasevic, Kim, Rashkin, Hajishirzi,
  Mehta, Bogar, Shevlin, Schütze, Yakura, Zhang, Wong, Ng, Noble, Jumelet,
  Geissinger, Kernion, Hilton, Lee, Fisac, Simon, Koppel, Zheng, Zou, Kocoń,
  Thompson, Kaplan, Radom, Sohl-Dickstein, Phang, Wei, Yosinski, Novikova,
  Bosscher, Marsh, Kim, Taal, Engel, Alabi, Xu, Song, Tang, Waweru, Burden,
  Miller, Balis, Berant, Frohberg, Rozen, Hernandez-Orallo, Boudeman, Jones,
  Tenenbaum, Rule, Chua, Kanclerz, Livescu, Krauth, Gopalakrishnan, Ignatyeva,
  Markert, Dhole, Gimpel, Omondi, Mathewson, Chiafullo, Shkaruta, Shridhar,
  McDonell, Richardson, Reynolds, Gao, Zhang, Dugan, Qin, Contreras-Ochando,
  Morency, Moschella, Lam, Noble, Schmidt, He, Colón, Metz, Şenel, Bosma,
  Sap, ter Hoeve, Farooqi, Faruqui, Mazeika, Baturan, Marelli, Maru, Quintana,
  Tolkiehn, Giulianelli, Lewis, Potthast, Leavitt, Hagen, Schubert,
  Baitemirova, Arnaud, McElrath, Yee, Cohen, Gu, Ivanitskiy, Starritt, Strube,
  Swędrowski, Bevilacqua, Yasunaga, Kale, Cain, Xu, Suzgun, Tiwari, Bansal,
  Aminnaseri, Geva, Gheini, T, Peng, Chi, Lee, Krakover, Cameron, Roberts,
  Doiron, Nangia, Deckers, Muennighoff, Keskar, Iyer, Constant, Fiedel, Wen,
  Zhang, Agha, Elbaghdadi, Levy, Evans, Casares, Doshi, Fung, Liang, Vicol,
  Alipoormolabashi, Liao, Liang, Chang, Eckersley, Htut, Hwang, Miłkowski,
  Patil, Pezeshkpour, Oli, Mei, Lyu, Chen, Banjade, Rudolph, Gabriel, Habacker,
  Delgado, Millière, Garg, Barnes, Saurous, Arakawa, Raymaekers, Frank,
  Sikand, Novak, Sitelew, LeBras, Liu, Jacobs, Zhang, Salakhutdinov, Chi, Lee,
  Stovall, Teehan, Yang, Singh, Mohammad, Anand, Dillavou, Shleifer, Wiseman,
  Gruetter, Bowman, Schoenholz, Han, Kwatra, Rous, Ghazarian, Ghosh, Casey,
  Bischoff, Gehrmann, Schuster, Sadeghi, Hamdan, Zhou, Srivastava, Shi, Singh,
  Asaadi, Gu, Pachchigar, Toshniwal, Upadhyay, Shyamolima, Debnath, Shakeri,
  Thormeyer, Melzi, Reddy, Makini, Lee, Torene, Hatwar, Dehaene, Divic, Ermon,
  Biderman, Lin, Prasad, Piantadosi, Shieber, Misherghi, Kiritchenko, Mishra,
  Linzen, Schuster, Li, Yu, Ali, Hashimoto, Wu, Desbordes, Rothschild, Phan,
  Wang, Nkinyili, Schick, Kornev, Telleen-Lawton, Tunduny, Gerstenberg, Chang,
  Neeraj, Khot, Shultz, Shaham, Misra, Demberg, Nyamai, Raunak, Ramasesh,
  Prabhu, Padmakumar, Srikumar, Fedus, Saunders, Zhang, Vossen, Ren, Tong,
  Zhao, Wu, Shen, Yaghoobzadeh, Lakretz, Song, Bahri, Choi, Yang, Hao, Chen,
  Belinkov, Hou, Hou, Bai, Seid, Zhao, Wang, Wang, Wang, and
  Wu]{srivastava2022imitation}
Aarohi Srivastava, Abhinav Rastogi, Abhishek Rao, Abu Awal~Md Shoeb, Abubakar
  Abid, Adam Fisch, Adam~R. Brown, Adam Santoro, Aditya Gupta, Adrià
  Garriga-Alonso, Agnieszka Kluska, Aitor Lewkowycz, Akshat Agarwal, Alethea
  Power, Alex Ray, Alex Warstadt, Alexander~W. Kocurek, Ali Safaya, Ali Tazarv,
  Alice Xiang, Alicia Parrish, Allen Nie, Aman Hussain, Amanda Askell, Amanda
  Dsouza, Ambrose Slone, Ameet Rahane, Anantharaman~S. Iyer, Anders Andreassen,
  Andrea Madotto, Andrea Santilli, Andreas Stuhlmüller, Andrew Dai, Andrew La,
  Andrew Lampinen, Andy Zou, Angela Jiang, Angelica Chen, Anh Vuong, Animesh
  Gupta, Anna Gottardi, Antonio Norelli, Anu Venkatesh, Arash Gholamidavoodi,
  Arfa Tabassum, Arul Menezes, Arun Kirubarajan, Asher Mullokandov, Ashish
  Sabharwal, Austin Herrick, Avia Efrat, Aykut Erdem, Ayla Karakaş, B.~Ryan
  Roberts, Bao~Sheng Loe, Barret Zoph, Bartłomiej Bojanowski, Batuhan Özyurt,
  Behnam Hedayatnia, Behnam Neyshabur, Benjamin Inden, Benno Stein, Berk
  Ekmekci, Bill~Yuchen Lin, Blake Howald, Cameron Diao, Cameron Dour, Catherine
  Stinson, Cedrick Argueta, César~Ferri Ramírez, Chandan Singh, Charles
  Rathkopf, Chenlin Meng, Chitta Baral, Chiyu Wu, Chris Callison-Burch, Chris
  Waites, Christian Voigt, Christopher~D. Manning, Christopher Potts, Cindy
  Ramirez, Clara~E. Rivera, Clemencia Siro, Colin Raffel, Courtney Ashcraft,
  Cristina Garbacea, Damien Sileo, Dan Garrette, Dan Hendrycks, Dan Kilman, Dan
  Roth, Daniel Freeman, Daniel Khashabi, Daniel Levy, Daniel~Moseguí
  González, Danielle Perszyk, Danny Hernandez, Danqi Chen, Daphne Ippolito,
  Dar Gilboa, David Dohan, David Drakard, David Jurgens, Debajyoti Datta, Deep
  Ganguli, Denis Emelin, Denis Kleyko, Deniz Yuret, Derek Chen, Derek Tam,
  Dieuwke Hupkes, Diganta Misra, Dilyar Buzan, Dimitri~Coelho Mollo, Diyi Yang,
  Dong-Ho Lee, Ekaterina Shutova, Ekin~Dogus Cubuk, Elad Segal, Eleanor
  Hagerman, Elizabeth Barnes, Elizabeth Donoway, Ellie Pavlick, Emanuele
  Rodola, Emma Lam, Eric Chu, Eric Tang, Erkut Erdem, Ernie Chang, Ethan~A.
  Chi, Ethan Dyer, Ethan Jerzak, Ethan Kim, Eunice~Engefu Manyasi, Evgenii
  Zheltonozhskii, Fanyue Xia, Fatemeh Siar, Fernando Martínez-Plumed,
  Francesca Happé, Francois Chollet, Frieda Rong, Gaurav Mishra, Genta~Indra
  Winata, Gerard de~Melo, Germán Kruszewski, Giambattista Parascandolo,
  Giorgio Mariani, Gloria Wang, Gonzalo Jaimovitch-López, Gregor Betz, Guy
  Gur-Ari, Hana Galijasevic, Hannah Kim, Hannah Rashkin, Hannaneh Hajishirzi,
  Harsh Mehta, Hayden Bogar, Henry Shevlin, Hinrich Schütze, Hiromu Yakura,
  Hongming Zhang, Hugh~Mee Wong, Ian Ng, Isaac Noble, Jaap Jumelet, Jack
  Geissinger, Jackson Kernion, Jacob Hilton, Jaehoon Lee, Jaime~Fernández
  Fisac, James~B. Simon, James Koppel, James Zheng, James Zou, Jan Kocoń, Jana
  Thompson, Jared Kaplan, Jarema Radom, Jascha Sohl-Dickstein, Jason Phang,
  Jason Wei, Jason Yosinski, Jekaterina Novikova, Jelle Bosscher, Jennifer
  Marsh, Jeremy Kim, Jeroen Taal, Jesse Engel, Jesujoba Alabi, Jiacheng Xu,
  Jiaming Song, Jillian Tang, Joan Waweru, John Burden, John Miller, John~U.
  Balis, Jonathan Berant, Jörg Frohberg, Jos Rozen, Jose Hernandez-Orallo,
  Joseph Boudeman, Joseph Jones, Joshua~B. Tenenbaum, Joshua~S. Rule, Joyce
  Chua, Kamil Kanclerz, Karen Livescu, Karl Krauth, Karthik Gopalakrishnan,
  Katerina Ignatyeva, Katja Markert, Kaustubh~D. Dhole, Kevin Gimpel, Kevin
  Omondi, Kory Mathewson, Kristen Chiafullo, Ksenia Shkaruta, Kumar Shridhar,
  Kyle McDonell, Kyle Richardson, Laria Reynolds, Leo Gao, Li~Zhang, Liam
  Dugan, Lianhui Qin, Lidia Contreras-Ochando, Louis-Philippe Morency, Luca
  Moschella, Lucas Lam, Lucy Noble, Ludwig Schmidt, Luheng He, Luis~Oliveros
  Colón, Luke Metz, Lütfi~Kerem Şenel, Maarten Bosma, Maarten Sap, Maartje
  ter Hoeve, Maheen Farooqi, Manaal Faruqui, Mantas Mazeika, Marco Baturan,
  Marco Marelli, Marco Maru, Maria Jose~Ramírez Quintana, Marie Tolkiehn,
  Mario Giulianelli, Martha Lewis, Martin Potthast, Matthew~L. Leavitt,
  Matthias Hagen, Mátyás Schubert, Medina~Orduna Baitemirova, Melody Arnaud,
  Melvin McElrath, Michael~A. Yee, Michael Cohen, Michael Gu, Michael
  Ivanitskiy, Michael Starritt, Michael Strube, Michał Swędrowski, Michele
  Bevilacqua, Michihiro Yasunaga, Mihir Kale, Mike Cain, Mimee Xu, Mirac
  Suzgun, Mo~Tiwari, Mohit Bansal, Moin Aminnaseri, Mor Geva, Mozhdeh Gheini,
  Mukund~Varma T, Nanyun Peng, Nathan Chi, Nayeon Lee, Neta Gur-Ari Krakover,
  Nicholas Cameron, Nicholas Roberts, Nick Doiron, Nikita Nangia, Niklas
  Deckers, Niklas Muennighoff, Nitish~Shirish Keskar, Niveditha~S. Iyer, Noah
  Constant, Noah Fiedel, Nuan Wen, Oliver Zhang, Omar Agha, Omar Elbaghdadi,
  Omer Levy, Owain Evans, Pablo Antonio~Moreno Casares, Parth Doshi, Pascale
  Fung, Paul~Pu Liang, Paul Vicol, Pegah Alipoormolabashi, Peiyuan Liao, Percy
  Liang, Peter Chang, Peter Eckersley, Phu~Mon Htut, Pinyu Hwang, Piotr
  Miłkowski, Piyush Patil, Pouya Pezeshkpour, Priti Oli, Qiaozhu Mei, Qing
  Lyu, Qinlang Chen, Rabin Banjade, Rachel~Etta Rudolph, Raefer Gabriel, Rahel
  Habacker, Ramón~Risco Delgado, Raphaël Millière, Rhythm Garg, Richard
  Barnes, Rif~A. Saurous, Riku Arakawa, Robbe Raymaekers, Robert Frank, Rohan
  Sikand, Roman Novak, Roman Sitelew, Ronan LeBras, Rosanne Liu, Rowan Jacobs,
  Rui Zhang, Ruslan Salakhutdinov, Ryan Chi, Ryan Lee, Ryan Stovall, Ryan
  Teehan, Rylan Yang, Sahib Singh, Saif~M. Mohammad, Sajant Anand, Sam
  Dillavou, Sam Shleifer, Sam Wiseman, Samuel Gruetter, Samuel~R. Bowman,
  Samuel~S. Schoenholz, Sanghyun Han, Sanjeev Kwatra, Sarah~A. Rous, Sarik
  Ghazarian, Sayan Ghosh, Sean Casey, Sebastian Bischoff, Sebastian Gehrmann,
  Sebastian Schuster, Sepideh Sadeghi, Shadi Hamdan, Sharon Zhou, Shashank
  Srivastava, Sherry Shi, Shikhar Singh, Shima Asaadi, Shixiang~Shane Gu, Shubh
  Pachchigar, Shubham Toshniwal, Shyam Upadhyay, Shyamolima, Debnath, Siamak
  Shakeri, Simon Thormeyer, Simone Melzi, Siva Reddy, Sneha~Priscilla Makini,
  Soo-Hwan Lee, Spencer Torene, Sriharsha Hatwar, Stanislas Dehaene, Stefan
  Divic, Stefano Ermon, Stella Biderman, Stephanie Lin, Stephen Prasad,
  Steven~T. Piantadosi, Stuart~M. Shieber, Summer Misherghi, Svetlana
  Kiritchenko, Swaroop Mishra, Tal Linzen, Tal Schuster, Tao Li, Tao Yu, Tariq
  Ali, Tatsu Hashimoto, Te-Lin Wu, Théo Desbordes, Theodore Rothschild, Thomas
  Phan, Tianle Wang, Tiberius Nkinyili, Timo Schick, Timofei Kornev, Timothy
  Telleen-Lawton, Titus Tunduny, Tobias Gerstenberg, Trenton Chang, Trishala
  Neeraj, Tushar Khot, Tyler Shultz, Uri Shaham, Vedant Misra, Vera Demberg,
  Victoria Nyamai, Vikas Raunak, Vinay Ramasesh, Vinay~Uday Prabhu, Vishakh
  Padmakumar, Vivek Srikumar, William Fedus, William Saunders, William Zhang,
  Wout Vossen, Xiang Ren, Xiaoyu Tong, Xinran Zhao, Xinyi Wu, Xudong Shen,
  Yadollah Yaghoobzadeh, Yair Lakretz, Yangqiu Song, Yasaman Bahri, Yejin Choi,
  Yichi Yang, Yiding Hao, Yifu Chen, Yonatan Belinkov, Yu~Hou, Yufang Hou,
  Yuntao Bai, Zachary Seid, Zhuoye Zhao, Zijian Wang, Zijie~J. Wang, Zirui
  Wang, and Ziyi Wu.
\newblock Beyond the imitation game: Quantifying and extrapolating the
  capabilities of language models, 2022.

\bibitem[Suzgun et~al.(2022)Suzgun, Scales, Scharli, Gehrmann, Tay, Chung,
  Chowdhery, Le, hsin Chi, Zhou, and Wei]{BBHSuzgun2022ChallengingBT}
Mirac Suzgun, Nathan Scales, Nathanael Scharli, Sebastian Gehrmann, Yi~Tay,
  Hyung~Won Chung, Aakanksha Chowdhery, Quoc~V. Le, Ed~Huai hsin Chi, Denny
  Zhou, and Jason Wei.
\newblock Challenging big-bench tasks and whether chain-of-thought can solve
  them.
\newblock \emph{ArXiv}, abs/2210.09261, 2022.

\bibitem[Chen et~al.(2021)Chen, Tworek, Jun, Yuan, Ponde, Kaplan, Edwards,
  Burda, Joseph, Brockman, Ray, Puri, Krueger, Petrov, Khlaaf, Sastry, Mishkin,
  Chan, Gray, Ryder, Pavlov, Power, Kaiser, Bavarian, Winter, Tillet, Such,
  Cummings, Plappert, Chantzis, Barnes, Herbert-Voss, Guss, Nichol, Babuschkin,
  Balaji, Jain, Carr, Leike, Achiam, Misra, Morikawa, Radford, Knight,
  Brundage, Murati, Mayer, Welinder, McGrew, Amodei, McCandlish, Sutskever, and
  Zaremba]{HumanEvalChen2021EvaluatingLL}
Mark Chen, Jerry Tworek, Heewoo Jun, Qiming Yuan, Henrique Ponde, Jared Kaplan,
  Harrison Edwards, Yura Burda, Nicholas Joseph, Greg Brockman, Alex Ray, Raul
  Puri, Gretchen Krueger, Michael Petrov, Heidy Khlaaf, Girish Sastry, Pamela
  Mishkin, Brooke Chan, Scott Gray, Nick Ryder, Mikhail Pavlov, Alethea Power,
  Lukasz Kaiser, Mohammad Bavarian, Clemens Winter, Philippe Tillet,
  Felipe~Petroski Such, David~W. Cummings, Matthias Plappert, Fotios Chantzis,
  Elizabeth Barnes, Ariel Herbert-Voss, William~H. Guss, Alex Nichol, Igor
  Babuschkin, S.~Arun Balaji, Shantanu Jain, Andrew Carr, Jan Leike, Joshua
  Achiam, Vedant Misra, Evan Morikawa, Alec Radford, Matthew~M. Knight, Miles
  Brundage, Mira Murati, Katie Mayer, Peter Welinder, Bob McGrew, Dario Amodei,
  Sam McCandlish, Ilya Sutskever, and Wojciech Zaremba.
\newblock Evaluating large language models trained on code.
\newblock \emph{ArXiv}, abs/2107.03374, 2021.

\bibitem[Askell et~al.(2021)Askell, Bai, Chen, Drain, Ganguli, Henighan, Jones,
  Joseph, Mann, DasSarma, Elhage, Hatfield-Dodds, Hernandez, Kernion, Ndousse,
  Olsson, Amodei, Brown, Clark, McCandlish, Olah, and
  Kaplan]{askell2021general}
Amanda Askell, Yuntao Bai, Anna Chen, Dawn Drain, Deep Ganguli, Tom Henighan,
  Andy Jones, Nicholas Joseph, Ben Mann, Nova DasSarma, Nelson Elhage, Zac
  Hatfield-Dodds, Danny Hernandez, Jackson Kernion, Kamal Ndousse, Catherine
  Olsson, Dario Amodei, Tom Brown, Jack Clark, Sam McCandlish, Chris Olah, and
  Jared Kaplan.
\newblock A general language assistant as a laboratory for alignment, 2021.

\bibitem[Zheng et~al.(2023)Zheng, Chiang, Sheng, Zhuang, Wu, Zhuang, Lin, Li,
  Li, Xing, Zhang, Gonzalez, and Stoica]{zheng2023judging}
Lianmin Zheng, Wei-Lin Chiang, Ying Sheng, Siyuan Zhuang, Zhanghao Wu, Yonghao
  Zhuang, Zi~Lin, Zhuohan Li, Dacheng Li, Eric.~P Xing, Hao Zhang, Joseph~E.
  Gonzalez, and Ion Stoica.
\newblock Judging llm-as-a-judge with mt-bench and chatbot arena, 2023.

\bibitem[Lv et~al.(2023)Lv, Yang, Liu, Gao, Guo, and Qiu]{lv2023parameter}
Kai Lv, Yuqing Yang, Tengxiao Liu, Qinghui Gao, Qipeng Guo, and Xipeng Qiu.
\newblock Full parameter fine-tuning for large language models with limited
  resources, 2023.

\end{thebibliography}
\bibliographystyle{unsrtnat}
\end{document}